%% file: pcsvb0.tex
\documentclass{article} 
\usepackage{nips13submit_e,times}
\usepackage{url}
\usepackage{graphicx}
\usepackage{algorithm,algorithmic}
\usepackage{amsmath,amsfonts}
\usepackage[numbers,sectionbib]{natbib}
\usepackage{bibhacks}
\usepackage{bbm}
\usepackage{multirow,multicol}
\usepackage{mdwlist}
\usepackage{framed}

\newcommand{\E}{\mathbb{E}}
\newcommand{\I}{\mathbbm{1}}

\title{Practical Collapsed Stochastic Variational Inference for the HDP}

\author{
Arnim Bleier\\
Knowledge Technologies for the Social Sciences\\
Leibniz Institute for the the Social Sciences\\
Cologne, 50667 - Germany \\
\texttt{arnim.bleier@gesis.org} \\
}

\nipsfinalcopy 

\begin{document}

	\maketitle

	\begin{abstract}
		Recent advances have made it feasible to apply the stochastic variational paradigm to a collapsed representation of latent Dirichlet allocation (LDA). While the stochastic variational paradigm has successfully been applied to an uncollapsed representation of the hierarchical Dirichlet process (HDP), no attempts to apply this type of inference in a collapsed setting of non-parametric topic modeling have been put forward so far. In this paper we explore such a collapsed stochastic variational Bayes inference for the HDP. The proposed online algorithm is easy to implement and accounts for the inference of hyper-parameters. First experiments show a promising improvement in predictive performance.
	\end{abstract}
	
	\input{background}

	\input{pcvb0}

	\input{stochastic_updates}

	\input{evaluation}

	\input{discussion}

	\bibliography{../references}{}

\end{document}

%% file: background.tex
\section{Background}

We begin by considering a model where each document d is a mixture $\theta_d$ of K discrete topic-distributions $\phi_k$ over a vocabulary of V terms. Let $z_{di} \in \{1,..,K\}$ denote the topic of the $i^{th}$ word $w_{di} \in \{1,..,V\}$  in document $d  \in \{1,..,D\}$ and place Dirichlet priors on the parameters $\theta_d$, $\phi_k$. We have 

\begin{align*}
	z_{di} \mid \theta_d &\sim Discrete(\theta_d) \mbox{ ,}  &&& 
	\theta_d &\sim Dirichlet(\alpha\pi)  \mbox{ ,} \\
	w_{di} \mid z_{di}, \{\phi_k\}  &\sim Discrete(\phi_{z_{di}}) \mbox{ ,} &&& 
	\phi_k &\sim Dirichlet(\beta) \mbox{ ,}
\end{align*}

where $\pi$ is the top-level distribution over topics, and $\alpha$ and $\beta$ are concentration parameters. While the dimensionality of K is fixed in latent Dirichlet allocation (LDA), we want the model to determine the number of topics needed. Consequently we follow the assumptions made by the hierarchical Dirichlet process (HDP) \cite{teh2006hierarchical} of a countable but infinite number of topics, of which only a finite number is used in the posterior. Our prior $\pi$ is constructed by a truncated sick-breaking process \cite{teh2007collapseddfs},

\begin{align}
	\label{eq:sick_breaking2}
	\pi_k = \bar{\pi}_k \prod_{l=1}^{k-1}(1-\bar{\pi}_l) \mbox{ ,} &&& 
	\bar{\pi}_k \mid \gamma \sim Beta(1,\gamma) \mbox{ ,} &&& 
	\bar{\pi}_T = 1 \mbox{ ,}
\end{align}

where $\bar{\pi}$ are the stick proportions. T is the truncation level and not the number of topics; if set to an appropriate level (i.e. $T > K$) the truncated stick-breaking process is a sufficient approximation of the Dirichlet process.

In the reminder of this paper we start by reviewing variational batch inference for the collapsed representation of the HDP. We then introduce our proposed stochastic updates. After that an early evaluation of our algorithm is presented. We conclude with a discussion of our work and its current limitations.

%% file: pcvb0.tex
\section{Practical Collapsed Variational Inference}

In this section we review practical batch collapsed variational Bayes inference (PCVB0) proposed by Sato et al. \citep{sato2012practical} which later will be the fundament of our stochastic inference. The collapsed representation of the HDP is achieved by marginalizing over $\theta$ and $\phi$. If only zero-order information is used, the update for the variational distributions $z_{di}$ over T possible topic assignments for each word is given by

\begin{equation}
	\label{eq:cvb0}
	q(z_{di} = k) \propto (n_{dk}^{\neg di} + \alpha\pi_k) 
	\frac{n_{k w_{di}}^{\neg di} + \beta}{n_{k.}^{\neg di} + V\beta}  \mbox{ .}
\end{equation}

Where $n_{dk}^{\neg di}$ is the number of times a word in document d has been assigned to topic k, and $n_{k w_{di}}^{\neg di}$ is the number of times the term $w_{di}$ has been assigned to topic k, in both cases excluding the current word $di$. Furthermore . is used in place of a variable to indicate that the sum over its values (i.e. $n_{k.} = \sum_w n_{kw}$) is taken. Teh's \citep{teh2007collapseddfs} original variational Bayes inference required maintaining variance counts for $\alpha\pi$. In a response Sato et al. \citep{sato2012practical} showed the usefulness of a lower-bound approximation for the number of tables in the Dirichlet process Chinese Restaurant representation. Leading to the update of the corpus-wide topic popularity

\begin{align}
	\pi_k &= \bar{\pi}_k \prod_{l=1}^{k-1}(1-\bar{\pi}_l) \mbox{ ,} \label{eq:hyper_pi} &&& 
	\bar{\pi}_k &= \frac{u_k}{u_k + v_k} \mbox{ ,} \\
	u_k &= 1 +  \sum_d \E[\I(n_{dk} \geq 1 )] \mbox{ ,} &&&
	v_k &= \gamma_0 + \sum_{l=k+1,d}^T \E[\I(n_{dl} \geq 1 )] \mbox{ ,}
\end{align}
	
where $\I(.)$ is the indicator function and the expectation $\E[\I(n_{dk} \geq 1 )] = 1- \prod_{i}(1-q(z_{di}=k))$. Moreover, using point estimates the updates for the hyper-parameters $\alpha$ and $\gamma$ are
	
\begin{align}
	\alpha = \frac{\sum_{d,k}  \E[\I(n_{dk} \geq 1 )] } {\sum_d [\Psi(n_d + \alpha^{old}) - \Psi(\alpha^{old})]} \mbox{ ,} \label{eq:update_hyper_alpha}  \\
	\gamma = \frac{T-1}{\sum_{k=1}^{T-1} [\Psi(u_k + v_k) - \Psi(v_k)] } \mbox{ ,} \label{eq:update_hyper_gamma}
\end{align}

with $n_d$ being the number of words in document d and $\Psi(.)$ the digamma function.

%% file: stochastic_updates.tex
\section{Proposed Updates}

One of the main drawbacks of batch collapsed variational inference for the HDP are its high memory requirements. We propose to circumvent this. Following the ideas behind stochastic collapsed variational Bayesian inference (SCVB0) proposed by Foulds et al. \citep{foulds2013stochastic} we potentially allow for data to arrive in a stream, but maintain the simplicity of the original PCVB0 schema.

The practical collapsed stochastic variational Bayes inference for the hierarchical Dirichlet process (PCSVB0), we propose, processes one word at a time, serially processing each word from all documents in turn. Suppose we have a guess of the current PCVB0 statistics. Next, we draw the $i^{th}$ word $w_{di}$ of document d and compute its corresponding $z_{di}$ via Equation~\eqref{eq:cvb0}. The expected number of times k appears in the document $n_{dk}$, with respect to the current word, is $n_d  q(z_{di} = k)$. Furthermore, the expected number of times $n_{kw}$ topic k is used by term $w_{di}$ is $n q(z_{di} = k)$ and zero for all other terms, with n being the total number of words in the corpus. As we process word by word we compute new $z_{di}$'s, but do not store them. Consequently, we cannot subtract the current word in Equation~\eqref{eq:cvb0} and approximate the distribution by
	
\begin{equation}
	\label{eq:scvb0}
	q(z_{di} = k) \propto (n_{dk} + \alpha\pi_k) 
	\frac{n_{k w_{di}} + \beta}{n_{k.} + V\beta} \mbox{ .}
\end{equation}

With the variational distribution $z_{di}$ of the current word we are able to update the expected statistics for $n_{dk}$ and $n_{kw}$ via
	
\begin{align}
	n_{dk} &\leftarrow (1 - \rho_t^d) n_{dk} + \rho_t^d n_d  q(z_{di} = k) \label{eq:stochastic_update_n_dk} \mbox{ ,} \\
	n_{kw} &\leftarrow (1 - \rho_t^c) n_{kw} + \rho_t^c n  q(z_{di} = k) \I[w_{di}=w] \mbox{ ,} \label{eq:stochastic_update_n_kw} 
\end{align}

and $\rho_t$ is the step-size in update t. The parameters u and v for the the stick-breaking proportions required in the computation of Equation~\eqref{eq:hyper_pi} and Equation~\eqref{eq:update_hyper_gamma} are updated after a document is  processed. We re-order the sticks according to their sizes. The updates are

\begin{align}
	u_k &\leftarrow (1 -\rho_t^h) u_k + \rho_t^h (1+D \E[\I(n_{dk} \geq 1 )]) \label{eq:stochastic_update_hyper_u} \mbox{ ,}  \\
	v_k &\leftarrow (1 -\rho_t^h) v_k + \rho_t^h (\gamma+D \sum_{l=k+1}^{T} \E[\I(n_{dl} \geq 1 )] ) \label{eq:stochastic_update_hyper_v}  \mbox{ .} 
\end{align}

For the stochastic update of $\alpha$ we again assume that our entire corpus consists of the single document d repeated D times, leading to

\begin{equation}
	\label{eq:update_alpha_stochastic}
	\alpha \leftarrow (1 -\rho_t^h) \alpha + \rho_t^h
	\left(\frac{\sum_{k=1}^T  \E[\I(n_{dk} \geq 1 )] }
	{\Psi(n_d + \alpha) - \Psi(\alpha)}\right)  \mbox{ .} 
\end{equation}

This suggests an iterative procedure, altering between approximating $z_{di}$ and updating the expected count statistics word by word, and updating the global topic popularity along the hyper-parameters document by document.

\begin{algorithm}
	\begin{algorithmic}[1]
		\caption{\label{alg:pcsvb0} PCSVB0 HDP inference}
		\STATE Initialize $n_{kw}$, $n_{dk}$, $\pi$, $\alpha$, $\gamma$.\\
		\STATE Set step-size schedule for $\rho_t^d$, $\rho_t^c$ and $\rho_t^h$.\\
		\REPEAT
		\FOR{each document $d$}
		\FOR{each word $i$ in $d$}
		\STATE Compute $q(z_{di} = k)$ \mbox{  }(Equation \ref{eq:scvb0}).
		\STATE Update $n_{dk}$ \mbox{  }(Equation \ref{eq:stochastic_update_n_dk}).
		\STATE Update $n_{kw}$ \mbox{  }(Equation \ref{eq:stochastic_update_n_kw}).
		\ENDFOR
		\STATE Update $\pi_k$ \mbox{  }(Equation \ref{eq:hyper_pi} with \ref{eq:stochastic_update_hyper_u},\ref{eq:stochastic_update_hyper_v}).
		\STATE Update $\gamma$ and $\alpha$\mbox{  }(Equation \ref{eq:update_hyper_gamma},\ref{eq:update_alpha_stochastic}).
		\ENDFOR
		\UNTIL{stopping criterion is met.}
	\end{algorithmic}
\end{algorithm}

%% file: evaluation.tex
\section{Evaluation}

In this section we describe a first experimental analysis of the proposed PCSVB0 inference. We studied the predictive performance of the algorithm on the Associated Press (TREC-1) data. The dataset contains 398k tokens across 2250 documents, with a vocabulary size of 10932 unique terms. For the evaluation we compared the perplexity versus the number of documents seen for PCSVB0, SCVB0 and PCVB0. We trained the model on 80\% of the documents. All held out documents were split; 70\% of the tokens in each held out document were used to estimate the document parameters, the remaining 30\% were used to compute the perplexity. 

We used the step-size schedule $\rho_t = \frac{s}{(\tau+t)^{0.9}}$. For the update of $n_{kw}$ in iteration t the schedule $\rho_t^c$ was parameterized with $s = 10$ and $\tau = 1000$; the schedule for $n_{dk}$ $\rho_t^d$ was parameterized with $s = 1$ and $\tau = 10$; and the schedule for the global stick-breaking weights and hyper-parameter updates $\rho_t^h$ was parameterized with $s = 5$ and $\tau = 100$. The prior on the topic-simplex $\beta$ was set for all algorithms to $\beta = 0.01$. The prior on the document-simplex for SCVB0 was set to $\alpha = 0.1$. Furthermore, for PCSVB0 as well as PCVB0 a truncation-level of $T=200$ was used.

\begin{figure}[H]
	\centering
	\includegraphics[scale=.45]{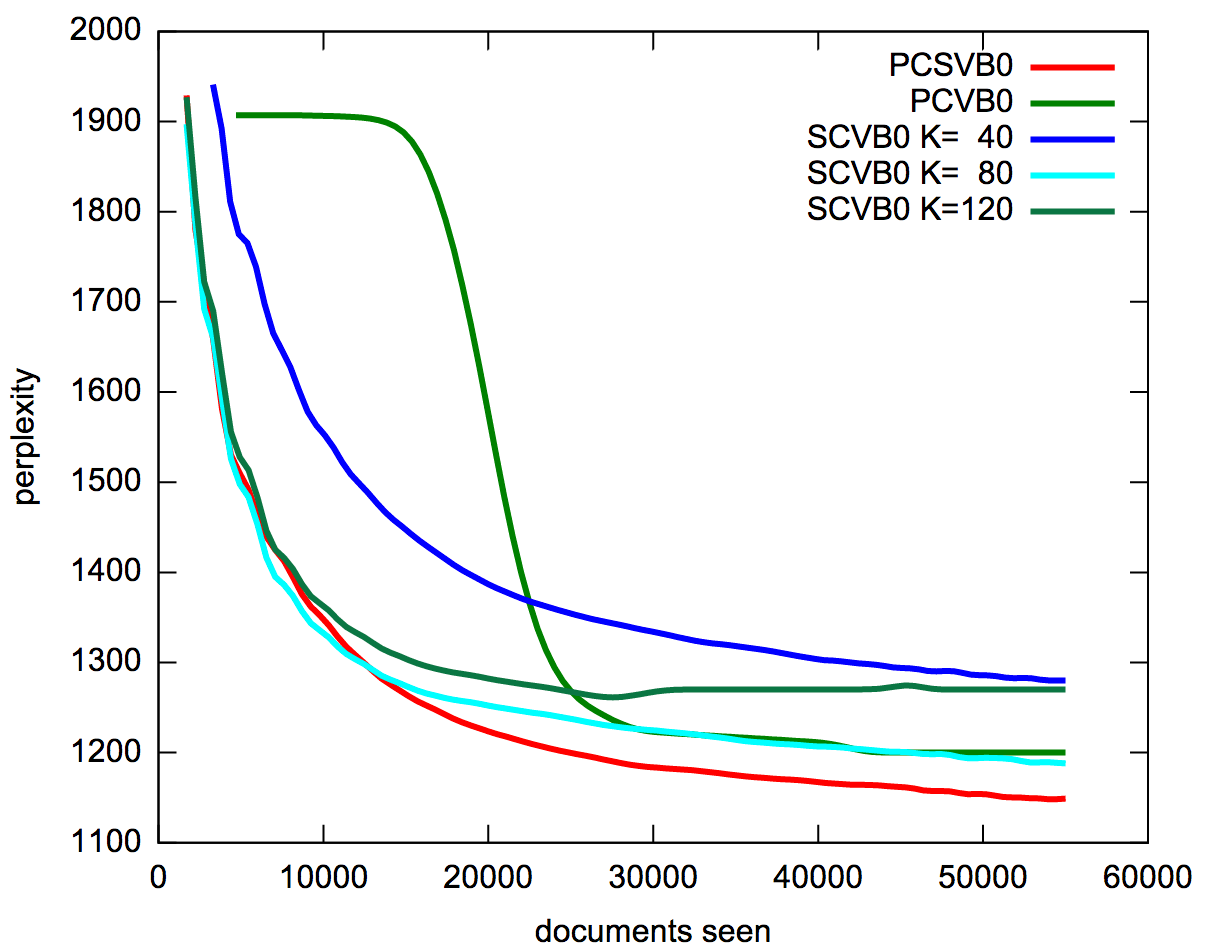}
	\caption{Comparison of predictive performance over 30 iterations on the TREC-1 dataset.}
\end{figure}

%% file: discussion.tex
\section{Discussion}

We presented a collapsed stochastic variational inference algorithm for the HDP-LDA topic model. Our algorithm is based on the application of a practical lower bound approximation of the truncated stick-breaking process to a collapsed stochastic inference scheme and simpler to implement then other uncollapsed online variational inference algorithms for the HDP\citep{wang2011online}. Initial small-scale experiments show promising improvements of PCSVB0 in predictive performance over existing algorithms, both in terms of the rate of convergence and the found optimum. Directions for future work are the application of so called `clumping' in order to perform the update only for each distinct term per document, and then to scale the update by the number of its copies. Another direction is the usage of mini-batches. Such optimizations would improve wall-clock time per iteration and allow for a fair comparison with other online variational inference algorithms for the HDP.